\documentclass[journal,compsoc]{IEEEtran}
\usepackage{ifluatex}
\ifluatex
\usepackage{unicode-math}
\fi

\usepackage{graphicx}
\usepackage{subfig}
\usepackage{lineno,hyperref}
\usepackage{amsmath}
\usepackage{float}
\usepackage{pgfplots}
\usepackage{csquotes}
\usepackage{graphicx}
\usepackage{subfig}
\usepackage{pgfplots}
\usepackage{float}
\usepackage{multirow}
\usepackage{xcolor}
\usepackage[linesnumbered,commentsnumbered,ruled,vlined]{algorithm2e}
\floatstyle{plaintop}
\restylefloat{table}
\ifCLASSOPTIONcompsoc
\usepackage[noadjust]{cite}
\else
\usepackage{cite}
\fi
\ifCLASSINFOpdf
\else
\fi
\begin{document}
	\title{X-CapsNet For Fake News Detection }
	
	\author{Mohammad~Hadi~Goldani,
		Reza~Safabakhsh,
		and~Saeedeh~Momtazi% <-this % stops a space
		\IEEEcompsocitemizethanks{\IEEEcompsocthanksitem M.H Goldani, R. Safabakhsh, and S. momtazi are with the Department
			of Computer Engineering, Amirkabir University of Technology, Tehran.\protect\\
			E-mail: goldani@aut.ac.ir, Safa@aut.ac.ir, momtazi@aut.ac.ir
		
			%\IEEEcompsocthanksitem J. Doe and J. Doe are with Anonymous University.
		}% <-this % stops an unwanted space
		\thanks{(Corresponding author: Reza safabakhsh)

			}}
	

	\markboth{}%
	{Shell \MakeLowercase{\textit{et al.}}: Bare Demo of IEEEtran.cls for Computer Society Journals}



	%\maketitle
	\IEEEtitleabstractindextext{%
		\begin{abstract}
			%The abstract goes here.
		News consumption has significantly increased with the growing popularity and use of web-based forums and social media. This sets the stage for misinforming and confusing people. To help reduce the impact of misinformation on users' potential health-related decisions and other intents, it is desired to have machine learning models to detect and combat fake news automatically. This paper proposes a novel transformer-based model using Capsule neural Networks(CapsNet) called X-CapsNet. This model includes a CapsNet with dynamic routing algorithm paralyzed with a size-based classifier for detecting short and long fake news statements. We use two size-based classifiers, a Deep Convolutional Neural Network (DCNN) for detecting long fake news statements and a Multi-Layer Perceptron (MLP) for detecting short news statements. To resolve the problem of representing short news statements, we use indirect features of news created by concatenating the vector of news speaker profiles and a vector of polarity, sentiment, and counting words of news statements. For evaluating the proposed architecture, we use the Covid-19 and the Liar datasets. The results in terms of the F1-score for the Covid-19 dataset and accuracy for the Liar dataset show that models perform better than the state-of-the-art baselines.
		\end{abstract}
		
}

	\maketitle

	%\IEEEdisplaynontitleabstractindextext

	% \fi
	\IEEEpeerreviewmaketitle

	%\IEEEraisesectionheading{\section{Introduction}\label{sec:introduction}}
	\section{Introduction}\label{sec:introduction}

	\IEEEPARstart{I}{n} recent years, online social media have become a common platform for broadcasting news for political, commercial, and entertainment purposes. News is understood as any information intended to make the public aware of the events happening around them, which may affect them personally or socially \cite{ verma2022mcred}.
	
	People use social media to search for and consume news due to its ease, convenience, and rapid spread \cite{ zhang2019overview}. These platforms have brought both constructive and destructive impacts. Therefore, as an integral part of culture and society, social media is a double-edged sword \cite{ ma2022dc}.
	People may manipulate and spread factual information for profit, or their entertainment in the form of fake news \cite{ bondielli2019survey}. Fake news played a pivotal role in the 2016 United States presidential election campaign after the mass of false information leaked on Facebook over the last three months of the presidential election \cite{ allcott2017social}.
	
	Misleading information can disrupt countries' economies, reduce people's trust in their governments, or promote a specific product to make huge profits. For example, this has already happened with COVID-19. Misleading information about lockdowns, vaccinations, and death statistics have fueled panic over purchasing groceries, disinfectants, masks, and paper products. This has led to shortages that have disrupted the supply chain and exacerbated the gaps between supply and demand and food insecurity \cite{elhadad2020detecting}. In addition, it caused a sharp decline in the international economy, severe losses in the value of crude oil, and the collapse of world stock markets \cite{ mhalla2020impact,albulescu2020coronavirus, gormsen2020coronavirus}. Furthermore, due to the spread of COVID-19 and the shortage of medical protection products worldwide, many people have lost faith in their governments, such as Italy and Iran \cite{ ling2020effects, ren2020fear}. These all drive the world into an economic recession \cite{ gormsen2020coronavirus,baldwin2020economics, sulkowski2020covid}.

	While a growing percentage of the population relies on social media platforms for news consumption, the reliability of the information shared remains an open issue.
	Fake news and many types of disinformation are rampant on social media, putting audiences around the world at risk. Therefore, detecting and mitigating the effects of disinformation is a crucial concern in studies where various approaches have been proposed, from linguistic indicators to deep learning models~\cite{bal2020analysing,Memon2020}.
	
	Recently, the automatic detection of fake news is attracting a large number of researchers \cite{ shu2019fakenewstracker, ghosh2018towards, PerezRosas:2018}. Early fake news detection methods often designed complete sets of hand-crafted features based on news content, user profiles, and news propagation paths, then train classifiers to discriminate the truthfulness of the news \cite{ castillo2011information, feng2012syntactic, yang2012automatic}. However, it is challenging to design all-encompassing features, as fake news is usually created on different writing styles, types of topics, and social media platforms \cite{ Shu:2017}. Therefore, many approaches based on deep neural networks \cite{ sahoo2021multiple, sedik2022deep, karimi2018multi, davoudi2022dss, chen2022using, zhang2018fake} have been proposed to automatically learn patterns discriminated by propagation paths and news content \cite{liao2021integrated}.
	
	The recent deep learning models improve the performance of fake news detection models, but the performance drops dramatically when the news content is short~\cite{ goldani2021detecting, goldani2021convolutional}. As a solution, in this work, we propose a new model based on CapsNet and indirect features for detecting fake news. The DCNN and MLP models with different feature extraction sections can be parallelized with a CapsNet architecture that is enhanced by using margin loss as the loss function. We also compare varieties of word representation layers and finally use the Bidirectional Encoder Representations from Transformers (BERT) \cite{kenton2019bert} and a robustly optimized BERT pretraining approach(RoBERTa) \cite{liu2019roberta} in our proposed model. We show that the proposed models achieve better results than the state-of-the-art methods on the Covid-19 and Liar fake news datasets.

	The rest of the paper is organized as follows: Section  \ref{sec:Section 2} reviews related work about fake news detection. Section \ref{sec:Section 3} presents the model proposed in this paper. The datasets used for fake news detection and evaluation metrics are introduced in Section \ref{sec:Section 4}. Section \ref{sec:Section 5} reports the experimental results, comparison with the baseline classification, and discussion. Finally, Section \ref{sec:Section 6} summarizes the paper.
	%I wish you the best of success.
	
	
	\section{Related work}
	\label{sec:Section 2}
	Social media have presently become the main source of information and news dissemination. This increases the challenges of spreading fake news. As a result, the identification of disinformation has been extensively studied in recent years, with the introduction of several tasks in the field. In this research, our focus is on the detection of fake news in two different domains, COVID-19 and politics, and is mainly based on the supervised method. A machine/deep learning model is trained based on the available data containing fake and real news. The model is then used to decide on the new news articles to find out if they are false or not. Most of the available studies investigating the fake news detection task have been conducted based on deep neural models, including CNN, Long Short Term Memory (LSTM), and BERT, which will be described in this section. The related works of the two domains are reviewed separately in this section.
	
	\subsection{COVID-19 domain}
	Since the appearance of the first case of COVID-19 on December 31, 2019, the World Health Organization (WHO) has declared Covid-19 as a pandemic emergency.
	As a source of COVID-19 information, tweets and social media news contain information or misinformation about COVID-19. For example, ordinary people become more eager to read more to know how to protect themselves \cite{abdelminaam2021coaid}.
	Brennen et al. \cite{brennen2020types} examined the sources of misinformation about COVID-19. Their analysis revealed that most of the COVID-19  misinformation is fabricated from real information rather than invented. Therefore the detection of COVID-19 fake news has attracted data scientists.

	Patwa et al. \cite{patwa2021fighting} provided a comprehensive dataset, the Covid-19 dataset, which includes fake and real news from Twitter. It includes 10,700 posts of the COVID-19 outbreak shared on social media. The real news was captured from 14 official Twitter accounts, and fake data were collected from social media and fact-checking websites. They performed a binary classification task (real vs. false) and evaluated four baselines of machine learning, namely Decision Tree (DT), Support Vector Machine (SVM), Logistic Regression (LR), and Gradient Boosting Decision Tree (GDBT). They achieved the best performance of 93.32 \% F1-score with SVM on the test set.
	
	Shifath et al. \cite{shifath2021transformer} used eight different pre-trained transformer-based models with additional layers to build a stackable ensemble classifier and refined them for the proposed models. Their models were evaluated on the Covid-19 dataset and showed that the RoBERTa-CNN model achieves 96.49\% F1 score on the test dataset.
	
	Several supervised text classification algorithms were evaluated by Wani et al. \cite{wani2021evaluating} on the Covid-19 fake news detection dataset. Their models are based on CNN, LSTM, and BERT. They also assessed the importance of unsupervised learning in the form of a pre-trained language model and distributed word representations using an unlabeled corpus of COVID-19 tweets. They claimed that their model improved the fake news detection accuracy.
	
	Samadi et al. \cite{Samadi2021} implemented three different neural classifiers with text representation models like BERT, RoBERTa, Funnel Transformer, and GPT2. They used Single Layer Perceptron (SLP), MLP, and CNN and tried to connect them to various contextualized text representation models. They compared the models' results and discussed their advantages and disadvantages. Finally, to corroborate the effectiveness of their approach, they selected the best model and compared their results with the most advanced models. Furthermore, they added a Gaussian noise layer to the combination of a contextualized text representation model with the CNN classifier. They claimed that the Gaussian noise layer could prevent overfitting in the learning process and, as a result, learns better about the Covid-19 and other data sets.
	
	A two-stage automated pipeline model was developed by Vijjali et al. \cite{vijjali2020two} for COVID-19 fake news detection. They used a state-of-the-art machine learning model for fake news detection. The first model used a novel fact-checking algorithm that searches for the most relevant facts regarding user claims for specific COVID-19 claims. The second model determines a claimant's level of truth by calculating the textual meaning between the claim and facts retrieved from a manually created Covid-19 dataset. They evaluated a set of models based on the classic text-based features as more contextualized Transformer-based models. They found that for both stages, the model pipelines based on BERT and ALBERT yield the best results.
	
	Koloski et al. \cite {koloski2021identification} leveraged several approaches and techniques for detecting COVID-19 fake news. They created several handcrafted features that capture the statistical distribution of characters and words in tweets. They showed that possible spatial representations were learned by capturing potentially relevant patterns from collections of n-grams of characters and features based on the words found in the tweets. For the assessment, they used various BERT-based representations to capture contextual information and the differences between fake and real COVID-19 news. Finally, they proved that the distilBERT tokenizer performs best with an F1 score of 97.05\%.
	
	\subsection{Politic domain}
	Many approaches based on neural networks and deep learning models are used to detect fake news articles in the datasets provided for the political domain. Different approaches, including CNNs, RNNs, hybrid models as well as more recent models such as CapsNets were used for the task.
	Liar dataset was presented by Wang et al.  \cite{Wang:2017}. Then they proposed a model that used statements and metadata together as inputs, a CNN for extracting features from statements, and a BiLSTM (Bi-directional long short-term memory) network for extracting features from metadata. They demonstrated that their model significantly improved the accuracy.
	
	Long et al. \cite{long2017fake} proposed a model on the Liar dataset that incorporates speaker profiles as features, containing speaker position, party affiliation, title, and credit history, into an attention-based LSTM model. They used two ways to improve the model with speaker profiles; (1) considering them in the attention model; (2) incorporating them as additional input data. They demonstrated that this model improves the performance of the classifier on the Liar dataset.
	
	The event adversarial neural network model was proposed by Wang et al. \cite{Wang:2018}. This model includes three main components: (1) the multimodal feature extractor, which uses CNN as its main module, (2) the fake news detector, which is a fully connected layer with softmax activation (3) the event discriminator that uses two completely connected layers and aims to classify the news in one of the K events based on the representations of the first components.
	A model based on CapsNet was also proposed for detecting fake news by Goldani et al. \cite{ goldani2021detecting}. They applied different levels of n-grams and different embedding models to news items of various lengths. Four filters with 2,3,4, and 5 kernel sizes and convolutional n-gram layers with non-static embedding were used for long news statements. For short news, only a static embedding of two filters with kernel sizes of 3 and 5 was used. They showed that their model improves the accuracy of the state-of-the-art methods.
	
	Choudhary et al. \cite{choudhary2021linguistic} used a language model to represent the news text to detect fake news. This linguistic model extracts information related to the news text's syntax, meaning, and readability. Due to the fact that this language model is time-consuming, they used a hierarchical model of neural networks to represent the features. In order to evaluate, the results obtained from sequential neural networks were compared with other machine learning methods and models based on LSTM. It was shown that the model based on hybrid neural networks could have better results than other methods.
	
	Blackledge et al. \cite{blackledge2021transforming} used transformer-based models to investigate the ability of these models to detect fake news and the generalizability of the models to detect fake news with different topics and models. They showed that the models could not naturally recognize news based on opinion and suspicion. Therefore, they proposed a new method to remove such news articles in the first step and then categorize them. In this article, it is shown that the generalizability using the proposed two-step method is able to improve the accuracy of the transformer-based models.
	\section{X-CapsNet for fake news detection}
	\label{sec:Section 3}

\begin{figure}[h]
		\centering
		\includegraphics[scale=0.35]{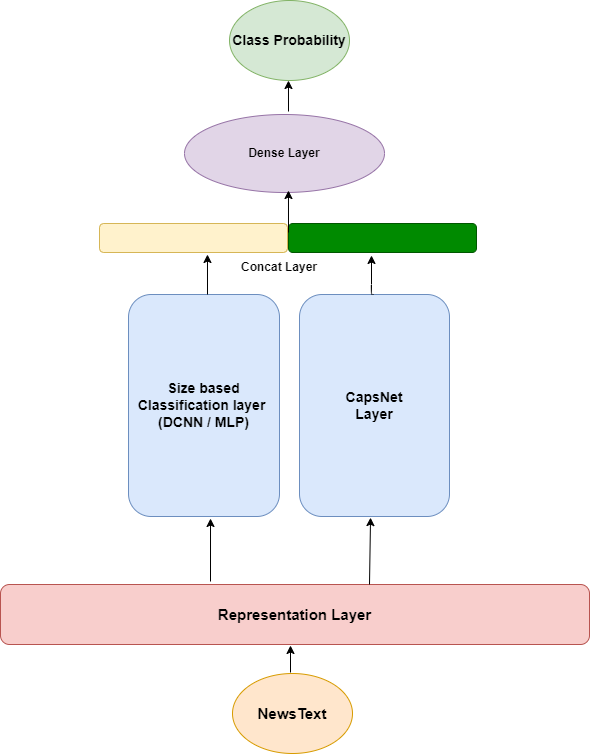}
		\caption{Proposed model for fake news detection}
		\label{fig:proposed}
	\end{figure}

	This section presents the models proposed for fake news detection in this paper. Depending on whether the news sentences are short or long, two different structures have been proposed to detect fake news. In these models, two parallel networks are concatenated: a CapsNet layer and a new size-based classification layer that uses a DCNN with pre-trained language models or an MLP layer with indirect features extracted from the input text. The concatenated layer is added to a dense layer to be used for detecting fake news eventually. Figure \ref{fig:proposed} shows the architecture of the proposed model.
	
	A CapsNet with a representation layer based on a pre-trained embedding model is used for all input text. In addition, when the news text is long, the DCNN model with three feature extractors with kernel sizes of 2, 3, and 4 is used. When the sentences are short, an MLP model that takes advantage of the indirect features of the news is used. In the following, we first review the pre-trained language models, including BERT with non-static embedding, which incrementally uptrains and updates the word embeddings in the training phase. Then we describe the classifiers used for the learning process.
	%In the following, each of these architectures is explained.
	
	%\subsection{DCNN-CapsNet for long news text}

	\subsection{Representation layer}
	\label{sec:REP}
	%\subsubsection{Direct features}
	Word embedding is one of the most widely used techniques in Natural Language Processing (NLP), and the goal is to learn a low-dimensional vector representation for a word in a text ~\cite{li2017learning}.
	The power of word embedding algorithms such as Word2Vec ~\cite{mikolov2013efficient}, FastText ~\cite{bojanowski2017enriching},  and GloVe ~\cite{ pennington2014glove} in capturing semantic and syntactic word relationships has been proven. This capability has facilitated various NLP tasks, such as aspect extraction, part-of-speech tagging, and sentiment analysis ~\cite{ khikmatullaev2019capsule}.
	The idea of distributional semantics states that words occurring in the same context tend to have similar meanings. In fact, word embeddings reveal hidden relationships between words that can be used during a training process. However, the static embedding techniques mentioned above only capture limited semantic information due to providing a unique static embedding vector for a word in different contexts. Therefore, in recent years, researchers have implemented various deep transformer-based word representation techniques that can take contextual information into account to generate embedding vectors for the same word in different contexts.
	
	BERT ~\cite{kenton2019bert} is an unsupervised deep model that uses the transformer architecture ~\cite{vaswani2017attention} and has been trained on a huge text corpus using two different scenarios: (1) the masked language model that learns the relationships between words by using their adjacency in a sentence, and (2) the next sentence prediction that learns relations between sentences.
	
	RoBERTa ~\cite{liu2019roberta} is similar to BERT with some hyperparameters tuning and modifications to its learning process. Liu et al. ~\cite{liu2019roberta} claimed that BERT was undertrained. Therefore, in addition to a larger dataset with longer sequences for training the model, they trained the model with longer batches. They also compared various alternative training approaches together and, as a result, claimed that for enhancing the performance of the learning process, the next sentence prediction loss can be removed.
	
	GPT2 is a large transformer-based language model trained on various texts taken from web pages on the Internet \cite{radford2019language}. The GPT2 architecture is similar to the OpenAI GPT model \cite{radford2018improving}, and is fine-tuned using the four tasks: text classification, text similarity, text consequence, and question answering.
	
	Funnel Transformer is an efficient encoder-decoder architecture that reduces the input features' resolution (length) using a pool operation and embeds them into a lower-dimensional vector \cite{dai2020funnel}. In this model, the decoder is an additional part of the architecture. It is used to simulate a masked language or, in the ELECTRA pre-learning task \cite{clark2020pre}, a new method of learning the self-supervised representation of a language.
	
	In fake news detection,  Goldani et al. \cite{ goldani2021detecting} showed that when the training data size is large enough, the model's performance can improve by using non-static embedding. Therefore, we also use a non-static setting in our model to update the text representation model during the training phase.
	The recent deep learning models improve the performance of fake news detection models, but the performance drops dramatically when the news content is short. To resolve this problem, we extract more features from the news in addition to the features extracted from the word embedding of sentences. The new features include the signs and information in the sentences \cite{shahi2021exploratory} along with the history of the speaker profile. More specifically, we use the following indirect features:
	\begin{itemize}
		\item Count of words (length of a news article)
		\item Count of unique words
		\item Count of letters
		\item Count of stop words
		\item Polarity score
		\item Subjectivity score
		\item History of the speaker profile
	\end{itemize}
	
	%\subsection{Parallel layer}
	\subsection{DCNN layer for long news article}
\begin{figure}[h]
		\centering
		\includegraphics[scale=0.35]{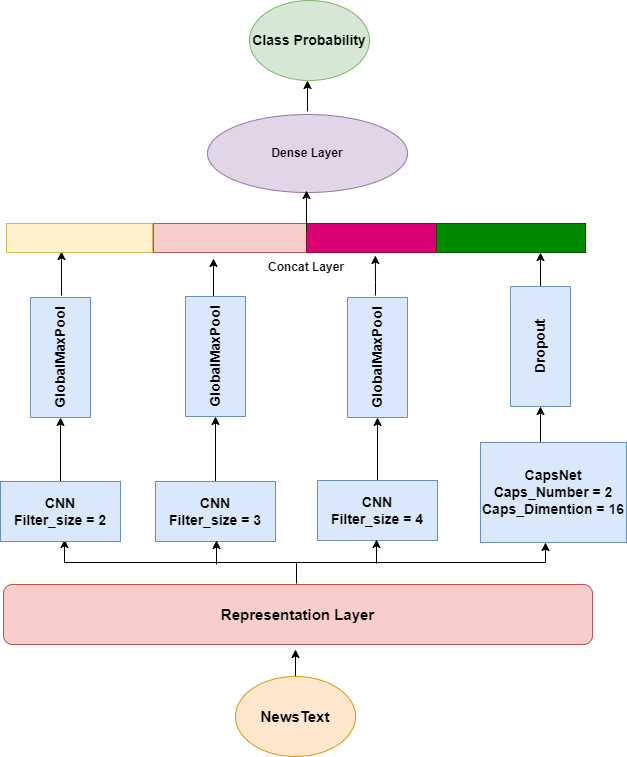}
		\caption{Proposed model for fake news detection in long news statements}
		\label{fig:proposed_model}
	\end{figure}

	In recent years, different variations of CNNs have been used in the task of fake news detection ~\cite{nasir2021fake, goldani2021convolutional, tan2021fn,kaliyar2020fndnet,wani2021evaluating}.
	A CNN architecture with convolutional and pooling layers can accurately extract features from local rendering to global rendering that indicate the powerful representational capabilities of CNNs.
	In order to extract more robust features for the learning process, the CNN needs to be enhanced with more identifying information. This requires that the intra-cluster similarity and inter-cluster dissimilarity of the learned features be maximized. For this goal, one of the most commonly used loss functions that are used with softmax in CNNs for fake news tasks is the margin loss \cite{goldani2021convolutional}. Using this loss function avoids overlapping problems and helps the model mitigate overfitting problems \cite{goldani2021convolutional}.
	In Figure \ref{fig:proposed_model}, the computational flow of the DCNN classifier is demonstrated. Zhong et al. \cite{zhong2019convolutional} showed that fake news detection could be investigated by adopting a standard text classification model consisting of an embedding layer, a one-dimensional convolutional layer, a max-pooling layer, and finally, a prediction-based output layer ~\cite{zhong2019convolutional}. Our proposed model is motivated by the concept of multiple parallel channels-variable-size-based neural networks considering three different filter sizes 2, 3, and 4 as n-gram convolutional layers for feature extraction.
	
	In this section, we present our fake news detection model for long news. The proposed model includes a pre-trained embedding model and two parallel classifiers. It reaps the benefits of both DCNN and CapsNet as two different neural network architectures that are used as classifiers. 
	
	Figure \ref{fig:proposed_model} shows the proposed model. In this architecture, four parallel neural networks have been used. These parallel networks include three different n-grams convolutional layers for feature extraction and a CapsNet layer that includes the primary capsule layer, a convolutional capsule layer, and a feed-forward capsule layer that was previously introduced by Yong et al. \cite{yang:2018}. Moreover, in the next layer, the outputs of CNNs and CapsNet go through a global max-pooling and a leaky-ReLU (Rectified Linear Unit) and concatenate. Then after using two dense layers, the final output predicts the label of the input news article. With this architecture, the models can learn more meaningful and extensive text representations on different n-gram levels according to the length of the text.

	\subsection{MLP layer for short news article}
\begin{figure}[h]
		\centering
		\includegraphics[scale=0.35]{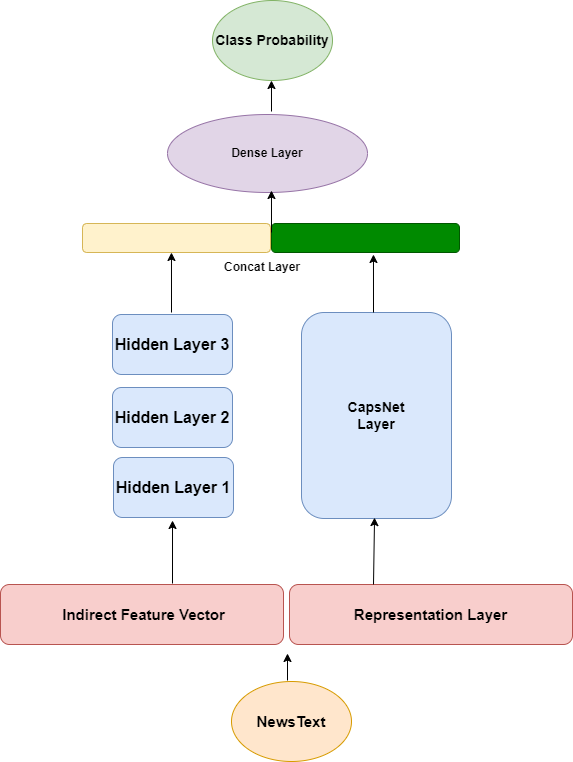}
		\caption{Proposed model for fake news detection in short news statements}
		\label{fig:proposed3}
	\end{figure}

	Due to the MLP's ability to learn, handle noisy or incomplete data and solve complex problems in real-time, this method is applied for the proposed classifier for the detection of short fake news \cite{ pahuja2021sound}. Figure \ref{fig:proposed3} shows the proposed model for detecting short news instances.
	
	In this model, indirect features are added in the training phase because of the short size of the news article and the need for more features for detecting fake news in addition to the transformer-based representation layer.
	
	The model is designed, implemented, and evaluated using the open-source Python software, and the Keras library. It consists of a fully connected 4-layer MLP Neural Network(NN) architecture with an input layer that takes data from the indirect characteristics of the input news text, two hidden layers to process the inputs, and an output layer that indicates the output. 
	
	Each layer is a computational abstraction consisting of a fixed number of computational data entities called neurons. Neurons are the building block of the NN that allow the NN to learn from the data and adjust its weights. Neurons in different layers are interconnected through weights that define the weight matrix between the layers.
	
	The model is designed sequentially by adding layers one after the other, starting with the input layer, the first hidden layer, the second hidden layer, and finally, the output layer. The number of input-outputs in the feature dataset determines the neurons in the NN input and output layer. The inner layers contain an arbitrary/specific number of neurons calculated empirically based on standard rules. Each level is defined by the number of nodes/neurons and the triggering function. The first layer has 12 input nodes corresponding to 12 feature attributes of the news indirect feature vector.
	
	The size of the input layer is 12×1. The hidden layer1 is the second layer. It is the first inner layer with 64 neurons in the optimized design and uses the ReLU activation function to operate on the inputs. ReLU is the most used activation function as it overcomes the problem of escape gradients during backpropagation and is suitable for large NNs. Each neuron in this layer is connected to all inputs of the weighted input layer. The size of the merge weight matrix of the input layer and the hidden layer is 12×64. The second hidden layer is the third layer. It has 32 neurons in the optimized form and uses the ReLU activation function. Each neuron in this layer is connected to the outputs of all neurons of the hidden layer1 with associated weights. The dimension of the matrix of the joining weights of hidden layer1 and hidden layer 2 is 32×32. The output layer is the fourth layer with 32 output nodes concatenated with the CapsNet output, and finally, a vector with 64 dimensions is fed to the dense layer.
	
	\subsection{CapsNet layer}
	
	After the success of CapsNets in various NLP tasks ~\cite{jiang2019challenge}, different models based on CapsNets have been used for fake news detection in recent years \cite{goldani2021detecting, palani2021cb, sridhar2021fake}.

	%%%%
	As mentioned in subsection\ref{sec:REP}, there are different ways for encoding a text. In this work, we use internal word embedding encoding, which means that in CapsNet, the input text is encoded. In this case, we use a 100-dimensional vector to represent a word with a batch size of 50. After evaluations, such dimensions prove to be sufficient to train CapsNet effectively.
	
	Considering the fake news detection with real and fake classes, low-level capsules should catch the most important words in the text. These are the words that significantly affect the classification. After that, two high-level capsules(real and fake classes) through dynamic routing detect the dependencies between significant words and recursively combine them to get the correct prediction. Algorithm \ref{alg:routing} shows the dynamic routing procedure, in which $r$ in STEP 2 in line 3 shows a hyperparameter that can be used for the training phase. In the "for loop", the scaler product $a_{ij} = v_j . u_{j|i}$ is defined as the agreement. The agreement is a log likelihood and is added to the initial logit, $b_{ij}$. The initial coefficients are refined iteratively. This operation measures the agreement between the output $v_j$ of each capsule $j$ in the above layer and the prediction $u_{j|i}$ that was made by capsule $i$.
	
	Consequently, the assumption is that low-level capsules will detect the words that significantly affect the classification of the text, and high-level capsules will detect low-level capsules and maximize the prediction value.
	
	In the original architecture of CapsNet \cite{Sabour:2017}, 2D convolution was used for image processing because the input is an image. It is important to consider the pixels because the surrounding pixels bring additional information. However, for fake news detection and in the case of text classification, it is not necessary to consider the surrounding pixels since it is not an image. A 1D convolution operation at the first and primary layers is used in this case.
	
	In the original architecture, the decoder framework reconstructs the input image and uses the decoder as a regularization method \cite{Sabour:2017}. However, in the text classification task, there is no reason to use the decoder because the task is only to classify the input into predefined categories. Therefore, the decoder structure is removed from the proposed architecture. Instead of using the decoder as the regularization method, the proposed models use a dropout layer against overfitting \cite{khikmatullaev2019capsule}.
	
	\IncMargin{1em}
	\begin{algorithm}[htbp]
		\SetKwData{Left}{left}
		\SetKwData{This}{this}
		\SetKwData{Up}{up}
		\SetKwFunction{Union}{Union}
		\SetKwFunction{FindCompress}{FindCompress}
		\SetKwInOut{Input}{Input}\SetKwInOut{Output}{Output}
		\SetKwComment{comment}{\#}{}
		\Input{$u_{i|j}$,$r$,$l$}
		\Output{$v_j$}
		\BlankLine
		
		\textbf{STEP 1:}  for all capsule $i$ in layer $l$ and capsule $j$ in layer ($l$+1): $b_{ij} \leftarrow 0$ \\
		\textbf{STEP 2:}  iterative routing: \\
		
		\For{$i$ in range ($r$)}{
			for all capsule $i$ in layer $l$: $c_i \leftarrow softmax(b_{ij}$)\\
			for all capsule $j$ in layer $l+1$: $s_j \leftarrow \Sigma_i c_{ij} u_{j|i}$\\
			for all capsule $j$ in layer $l+1$: $v_j \leftarrow squash(s_j)$\\
			for all capsule $i$ in layer $l$ and capsule j in layer $l$+1: $b_{ij} \leftarrow b_{ij} + u_{i|j} + v_j$\\
			%calculate $\Delta$A  = abs( a - b ) \\
			%\comment{a is something, b is another thing}
			%\If{ $\Delta$A  $\leq$ threshold}
			%{do  $\Delta$A  + c  }
			return $v_j$
		}	
		\caption{Dynamic Routing algorithm}
		\label{alg:routing}
	\end{algorithm}
	
\begin{figure}[h]
		\centering
		\includegraphics[scale=0.5]{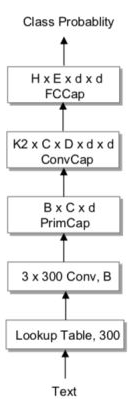}
		\caption{The architecture of capsule network proposed for text classification by ~\cite{yang:2018} }
		\label{fig:Yang}
	\end{figure}

	The CapsNet architecture includes a standard convolutional layer called an n-gram convolutional layer that acts as a feature extractor. The second layer maps the scalar features into a capsule representation called the primary capsule layer. The outputs of this capsule are fed to the new layer called the convolutional capsule layer. In this layer, each capsule is only connected to the local area of the layer below. In the final step, the previous layer's output is flattened and fed through the feed-forward capsule layer. For this layer, all capsules in the output are considered to be of a specific class. This architecture uses the maximum margin loss to train the model as presented in Figure \ref{fig:Yang} ~\cite{yang:2018}. 
	A CapsNet with two capsules of dimension 16 followed by a leaky-ReLU has been chosen as a parallelized neural network in the proposed model. 
	
	%%%%
	
	\subsection{Fully connected layer}
	The functionality of a dense layer is considered a linear operation in which all inputs are connected to all outputs by some weights. We use two dense layers to make the proposed model inherently dense. In the proposed model, the first dense layer takes the output of the concat layer, and then the second dense layer predicts the final output.

	\section{Evaluation}
	\label{sec:Section 4}
	The proposed model is evaluated in this section using different datasets for fake news detection.
	\subsection{Dataset}
	We use two datasets from different domains for the evaluation of the model. These datasets include the Covid-19 dataset and the Liar dataset.
	\subsubsection{The Covid-19 dataset}
	%\label{sec:Section 4.1.2}
	
	%\subsubsection{The Covid-19 dataset}
	When the COVID-19 pandemic began, social media users shared more and more misinformation and unconfirmed news about the Coronavirus. This motivated researchers to collect datasets from social media and propose machine learning models to evaluate their methods. The Covid-19 dataset is one of the recent datasets proposed by \cite{patwa2021fighting} that is a comprehensive dataset including fake and real news from Twitter. This dataset includes 10,700 posts about the COVID-19 outbreaks that were shared on social media. COVID-19 fake articles were collected from fact-checking websites and social media. Moreover, the real news was obtained from 14 official Twitter accounts.
	Table \ref{tab:Covid-19} shows the statistics of the dataset.
	
	\begin{table}[h]
		\centering
		\begin{tabular}{|c|cc|c|}
			\hline
			\multirow{2}{*}{\textbf{Dataset}} & \multicolumn{2}{c|}{\textbf{Label}}                & \multirow{2}{*}{\textbf{Total}} \\ \cline{2-3}
			& \multicolumn{1}{c|}{\textbf{Real}} & \textbf{Fake} &                                 \\ \hline
			Training                          & \multicolumn{1}{c|}{3360}          & 3360          & 4420                            \\ \hline
			Validation                        & \multicolumn{1}{c|}{1120}          & 1020          & 2140                            \\ \hline
			Test                              & \multicolumn{1}{c|}{1120}          & 1020          & 2140                            \\ \hline
		\end{tabular}
		\caption{Covid-19 dataset statistics by ~\cite{patwa2021fighting}}
		\label{tab:Covid-19}
	\end{table}
	
	\subsubsection{The Liar dataset}
	Another fake news dataset used for evaluating different models is the Liar dataset. This dataset contains 12,800 short political news texts from the United States in 6 different categories and is accessible from the \texttt{POLITIFACT.COM} website. Every news text has been validated on this site by a human agent. Therefore, the dataset is divided into 6 categories: true, false, mostly-True, half-true, barely-True \footnote{less true} and pants-fire\footnote{very false}. The distribution of labels is 1050 pants-fire labels, and the number of the other categories is between 2063 and 2638 \cite{Wang:2017}. For each news article, the metadata, such as speaker profiles, are taken into account in addition to news statements. This metadata includes valuable information about the new speaker's name, topic, job, state, party, and overall credit history. The total number of credit histories includes false counts, mostly-true counts, barely-true counts, half-true counts, and pants-fire counts. Table \ref{tab:LiarDataStatistics} demonstrates the statistics of the Liar dataset.

	\begin{table}[H]
		\centering
		\begin{tabular}{|lr|}
			\hline
			\multicolumn{2}{|l|}{Liar Dataset Statistics} \\\hline
			Training set size & 10,269\\
			Validation set size & 1,284\\
			Testing set size & 1,283\\
			Avg. statement length (tokens) & 17.9 \\\hline 
			\multicolumn{2}{|l|}{Top-3 Speaker Affiliations}\\ 
			Democrats & 4,150\\
			Republicans & 5,687\\
			None (e.g., FB posts) & 2,185\\\hline
			
		\end{tabular}
		\caption{The Liar dataset statistics provided by \cite{Wang:2017}}
		\label{tab:LiarDataStatistics}
	\end{table}
	
	\subsection{Evaluation metrics}
	In our experiments, the classification accuracy, precision, recall, and F1-score are used as evaluation metrics. The accuracy is  the ratio of correct predictions of the news label to the total number of news samples and is computed as:
	
	\begin{align}
	Accuracy = \frac{TP + TN}{TP + TN + FN + FP}
	\label{eq:2}
	\end{align}
	\\
	Precision shows the percentages of the reported fake news that are correctly detected:
	\begin{align}
	Precision = \frac{TP }{TP + FP}
	\label{eq:1}
	\end{align}
	\\
	Recall estimates the ratio of the correctly detected fake news:
	\begin{align}
	Recall = \frac{TP }{TP + FN}
	\label{eq:2}
	\end{align}
	\\
	F1-score is the harmonic mean of precision and recall:
	\begin{align}
	F1-score = 2 \times\frac{Precision+Recall}{Precision 2 \times Recall}
	\label{eq:3}
	\end{align}
	\\
	
	In these equations, TP represents the number of True Positive results, TN represents the number of True Negative results,  FP represents the number of False Positive results, and FN represents the number of False Negative results.

	\section{Results}
	\label{sec:Section 5}
	This section evaluates the proposed models on the Covid-19 and Liar datasets on different representation layers. The results are compared to other baseline methods, and the performance of parallel layers is evaluated separately. In the end, in the discussion subsection, a series of experiments on the dataset are discussed.

	\subsection{Classification results on the Covid-19 dataset}
	
	\subsubsection{Classification results on different representation on the Covid-19 dataset}
	% \usepackage{multirow}
	Table \ref{tab:RLCOVID} shows the evaluation of the proposed model on the Covid-19 dataset using different representation layers and routing iterations for the dynamic routing algorithm of the CapsNet. As it can be seen, the best result belongs to the model with RoBERTa as representation layer.
	\begin{table}[H]
		\centering
		\begin{tabular}{|c|c|c|c|c|c|}
			\hline
			\textbf{Model}                                                           & \textbf{RL} & \textbf{Acc}   & \textbf{Prec}  & \textbf{Rec}   & \textbf{F1}    \\ \hline
			\multirow{4}{*}{\begin{tabular}[c]{@{}c@{}}DCNN \\ CapsNet\end{tabular}} & BERT        & 96.77          & 96.55          & \textbf{97.48} & 97.00          \\ \cline{2-6} 
			& Funnel      & 97.05          & 96.85          & 97.20          & 97.02          \\ \cline{2-6} 
			& GPT2        & 97.00          & 96.63          & \textbf{97.48} & 97.05          \\ \cline{2-6} 
			& RoBERTa     & \textbf{97.34} & \textbf{97.21} & 97.38          & \textbf{97.29} \\ \hline
		\end{tabular}
		\caption{Classification results on different representations on the Covid-19 dataset}
		\label{tab:RLCOVID}
	\end{table}
	
	\subsubsection{Classification results on different routing iterations}
\begin{figure}[H]
		\pgfplotsset{
			% use this `compat' level or higher to use the advanced positioning of
			% the axis labels
			compat=1.5,
		}
		\begin{center}
			\begin{tikzpicture}
			
			\begin{axis}[
			xbar,
			y axis line style = { opacity = 0 },
			axis x line       = none,
			tickwidth         = 0pt,
			enlarge y limits  = 0.2,
			enlarge x limits  = 0.02,
			symbolic y coords = { RoBERTa, GPT2, Funnel, BERT},
			nodes near coords,
			typeset ticklabels with strut,
			%legend pos=outer north east ,
			legend style={at={(0.5,-0.15)},
				anchor=north,legend columns=-1},  
			]
			
			\addplot coordinates { (96.81,GPT2)
				(96.95,Funnel)  (96.73,BERT) (96.64,RoBERTa)};
			
			\addplot coordinates { (97.05,GPT2)
				(97.02,Funnel)  (97.00,BERT) (97.20,RoBERTa)};
			
			\addplot coordinates {  (96.97,GPT2)
				(96.64,Funnel)  (96.60,BERT) (97.29,RoBERTa) };
			
			%\legend pos=outer north east,                       
			\legend{ three iterations, two iterations, one iteration,}
			\end{axis}
			
			\end{tikzpicture}
		\end{center}
		
		\caption{Classification results on different routing iterations on the Covid-19 dataset} \label{tab:DRCOVID}
	\end{figure}
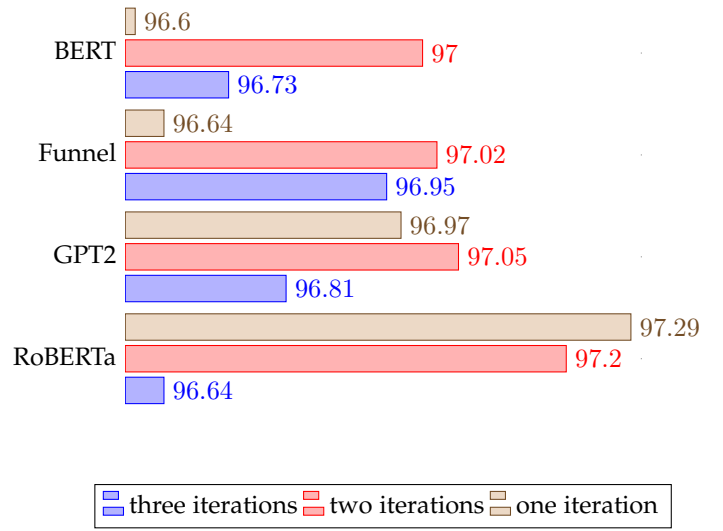

	Figure \ref{tab:DRCOVID} shows evaluations on the Covid-19 dataset in terms of the F1-score for different routing iterations for the dynamic routing algorithm of the CapsNet model. As a result, for long news statements with COVID-19 subjectivity, one repetition is sufficient to achieve the best result. This shows that combining higher hierarchies for data with more sentences is unnecessary to achieve better results, and the best results are obtained by repeating dynamic routing once.
	% \usepackage{multirow}

	\subsubsection{Classification results on the Covid-19 dataset}
	After the presentation of the Covid-19 dataset by ~\cite{patwa2021fighting}, different machine learning and deep learning models were evaluated for fake news detection on this dataset. ~\cite{ patwa2021fighting} used conventional machine learning models, including the DT, LR, SVM, and GDBT. ~\cite{shifath2021transformer} proposed an MLP connected to the RoBERTa’s pooled output by utilizing additional data for training. ~\cite{wani2021evaluating} evaluated many models, including a softmax layer connected to the BERT for prediction. ~\cite{Samadi2021} proposed a CNN connected to the RoBERTa’s pooled output and showed that the performance of this model is better than the previous models. Table \ref{tab:Result} compares our proposed model's results with the state-of-the-art models on the Covid-19 test set. The results in terms of the F1-score show that the DCNN-CapsNet model can perform better than the state-of-the-art baselines. Moreover, it should be mentioned that recall is an important factor in fake news detection since missing any fake news article has its own negative side effects. As can be seen in the tabulated results, we achieved the best recall among competitors.
	
	\begin{table}[h]
		\centering
		\begin{tabular}{|c|c|c|c|c|}
			\hline
			\textbf{Model}                       & \textbf{Acc} & \textbf{Prec} & \textbf{Rec} & \textbf{F1}    \\ \hline
			DT   ~\cite{patwa2021fighting}                                & 85.37             & 85.47              & 85.37           & 85.39          \\ \hline
			LR       ~\cite{patwa2021fighting}                            & 91.96             & 92.01              & 91.96           & 91.96          \\ \hline
			SVM       ~\cite{patwa2021fighting}                           & 93.32             & 93.33              & 93.32           & 93.32          \\ \hline
			GDBT     ~\cite{patwa2021fighting}                            & 86.96             & 87.24              & 86.96           & 86.96          \\ \hline
			RoBERTa-MLP     ~\cite{shifath2021transformer}               & 96.68             & 97.12              & 95.880          & 96.49          \\ \hline
			BERT-MLP          ~\cite{wani2021evaluating}         & 95.79             & \textbf{98.94}     & 92.15           & 95.43          \\ \hline
			RoBERTa-CNN           ~\cite{Samadi2021}               & \textbf{97.43}    & 98.30              & 96.27           & 97.27          \\ \hline
			Proposed Model & 97.34             & 97.21              & \textbf{97.38}  & \textbf{97.29} \\ \hline
		\end{tabular}
		\caption{Comparison of proposed model result with the result of other models on the Covid-19 test set.}
		\label{tab:Result}
	\end{table}
	
	\subsubsection{Performance of parallel layers}
	Table \ref{tab:Result2} shows the proposed model's performance compared to the two parallel models. The results show that using different feature extractors for CNN and adding CapsNet, which aims at keeping detailed information about the location of the object and its pose throughout the network, can improve the performance of the baseline models. The best result is achieved when both models are used together.
	
	\begin{table}[H]
		
		\begin{tabular}{|c|c|c|c|c|}
			\hline
			\textbf{Model} & \textbf{Acc} & \textbf{Prec} & \textbf{Rec} & \textbf{F1}    \\ \hline
			RoBERTa-CapsNet & 93.22             & 93.92              & 91.91  & 96.31 \\ \hline
			RoBERTa-CNN & \textbf{97.43}             & \textbf{98.30}              & 96.27  & 97.27 \\ \hline
			Proposed Model & 97.34             & 97.21              & \textbf{97.38}  & \textbf{97.29} \\ \hline
		\end{tabular}
		\caption{Comparison of proposed model result with the result of parallel layers on the Covid-19 test set.}
		\label{tab:Result2}
	\end{table}
	
	%%%%%%%%%%%%%%%%%%%%
	\subsection{Classification results on the Liar dataset}
	\subsubsection{Classification results on different representations on the Liar dataset}
	
	Table \ref{tab:RLIAR} shows the evaluation of the proposed model on the Liar dataset using different representation layers and routing iterations for the dynamic routing algorithm of the CapsNet. As it can be seen, the best result belongs to the model with RoBERTa as the representation layer.
	% \usepackage{multirow}
	\begin{table}[H]
		\begin{tabular}{|c|c|c|c|}
			\hline
			\textbf{Model}                                                          & \multicolumn{1}{l|}{\textbf{Representation}} & \textbf{Validation} & \textbf{Test}  \\ \hline
			\multirow{4}{*}{\begin{tabular}[c]{@{}c@{}}MLP \\ CapsNet\end{tabular}} & BERT                                         & 38.55               & 35.70          \\ \cline{2-4} 
			& Funnel                                       & 37.69               & 35.46          \\ \cline{2-4} 
			& GPT2                                         & 42.91               & 39.67          \\ \cline{2-4} 
			& RoBERTa                                      & \textbf{41.19}      & \textbf{41.77} \\ \hline
		\end{tabular}
		\caption{Classification results on different representations on the Liar dataset }
		\label{tab:RLIAR}
	\end{table}
	
	\subsubsection{Classification results on different routing iterations}
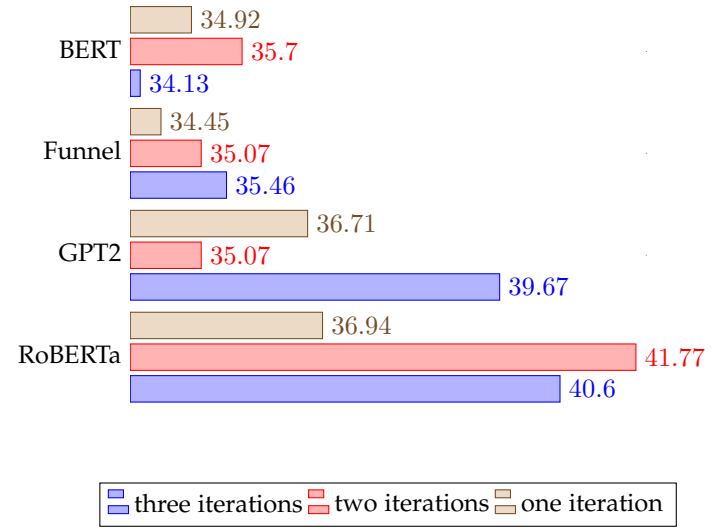
\begin{figure}[H]
		\pgfplotsset{
			% use this `compat' level or higher to use the advanced positioning of
			% the axis labels
			compat=1.5,
		}
		\begin{center}
			\begin{tikzpicture}
			
			\begin{axis}[
			xbar,
			y axis line style = { opacity = 0 },
			axis x line       = none,
			tickwidth         = 0pt,
			enlarge y limits  = 0.2,
			enlarge x limits  = 0.02,
			symbolic y coords = { RoBERTa, GPT2, Funnel, BERT},
			nodes near coords,
			typeset ticklabels with strut,
			%legend pos=outer north east ,
			legend style={at={(0.5,-0.15)},
				anchor=north,legend columns=-1},  
			]
			
			\addplot coordinates { (39.67,GPT2)
				(35.46,Funnel)  (34.13,BERT) (40.60,RoBERTa)};
			
			\addplot coordinates { (35.07,GPT2)
				(35.07,Funnel)  (35.70,BERT) (41.77,RoBERTa)};
			
			\addplot coordinates {  (36.71,GPT2)
				(34.45,Funnel)  (34.92,BERT) (36.94,RoBERTa) };
			
			%\legend pos=outer north east,                       
			\legend{ three iterations, two iterations, one iteration,}
			\end{axis}
			
			\end{tikzpicture}
		\end{center}
		
		\caption{Classification results on different routing iterations on the Liar dataset} \label{tab:DRLIAR}
	\end{figure}

	Figure \ref{tab:DRLIAR} shows evaluations on the Liar dataset in terms of the accuracy for different routing iterations for the dynamic routing algorithm of the CapsNet model. As a result, more repetition is needed for short news statements with political subjectivity to achieve the best result. This shows that combining higher hierarchies for data with more sentences is necessary to achieve better results, and the best results are obtained by repeating dynamic routing twice.
	
	% \usepackage{multirow}

	\subsubsection{Classification results on the Liar dataset}
	Table \ref{tab:RRLIAR} compares our proposed model's results with the state-of-the-art models on the Liar validation and test set. The results in terms of accuracy show that the MLP-CapsNet model can perform better than the state-of-the-art baselines. 
	
	\begin{table}[H]
		\begin{tabular}{|c|c|c|}
			\hline
			\textbf{Model}       & \textbf{Validation(\%)} & \textbf{Test(\%)}  \\ \hline
			Hybrid CNN~\cite{Wang:2017}          & 24.60               & 24.10          \\ \hline
			LSTM attention~\cite{long2017fake}       & 37.80               & 38.50          \\ \hline
			Two stage BERT model~\cite{liu2019two} & -                   & 40.58          \\ \hline
			CapsNet~\cite{goldani2021detecting}              & 40.90               & 39.50          \\ \hline
			CNN with margin loss~\cite{goldani2021convolutional} & \textbf{44.40}      & 41.60          \\ \hline
			Proposed model       & 41.19               & \textbf{41.77} \\ \hline
		\end{tabular}
		\caption{Comparison of proposed model result with the result of other models on the Liar validation and test set.}
		\label{tab:RRLIAR}
	\end{table}

	\subsubsection{Performance of parallel layers}
	Table \ref{tab:ResultL} shows the proposed model's performance compared to the two parallel models. The results show that using indirect features and adding CapsNet, which aims at keeping detailed information about the location of the object and its pose throughout the network, can improve the performance of the baseline models. The best result is achieved when both models are used together.
	
	\begin{table}[H]
		\centering
		\begin{tabular}{|c|c|c|}
			\hline
			\textbf{Model}       & \textbf{Validation(\%)} & \textbf{Test(\%)}  \\ \hline
			CapsNet              & 40.90               & 39.50          \\ \hline
			CNN with margin loss & \textbf{44.40}      & 41.60          \\ \hline
			Proposed model       & 41.19               & \textbf{41.77} \\ \hline
		\end{tabular}
		\caption{Comparison of proposed model result with the result of parallel layers on the Liar validation and test set.}
		\label{tab:ResultL}
	\end{table}
	\subsection{Discussion}
	
\begin{figure}[H]
		\centering
		\includegraphics[scale=0.45]{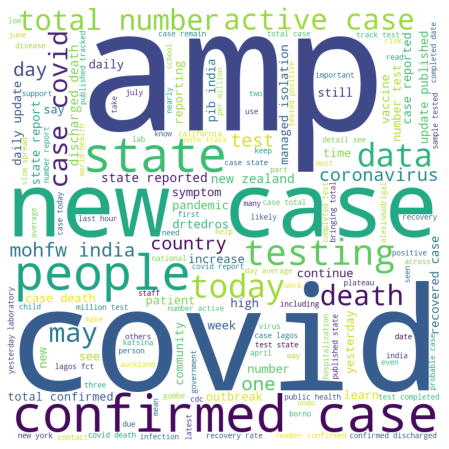}
		\caption{Word cloud of real news}
		\label{fig:d1}
	\end{figure}

\begin{figure}[H]
		\centering
		\includegraphics[scale=0.35]{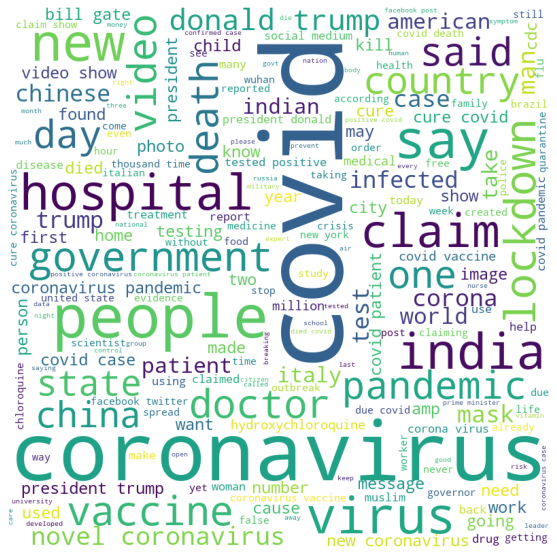}
		\caption{Word cloud of fake news}
		\label{fig:d2}
	\end{figure}

	%\subsubsection{Covid-19 dataset}
	This section further analyzes the training set of the Covid-19 dataset for real and fake news labels. Figures \ref{fig:d1} and \ref{fig:d2} show the word clouds for the real and fake news of the training set after omitting the stopwords, respectively. From the word clouds and most frequent words, we see an overlap of the important words across fake and real news. Therefore for more analysis, we list the ten most frequent words in real and fake news after removing the stopwords:
	\begin{itemize}
		\item \textbf{Fake news}: covid, Coronavirus, people, claim, trump, virus, say, vaccine, new, and case 
		
		\item \textbf{Real news}: case, covid, new, state, test, number, death, India, total, day
	\end{itemize}
	If we ignore the common words of the two groups, we find that among the fake news, words about sensitive quotes and reports such as the vaccine, Coronavirus, and the names of politicians are more frequent. Also, statistics about the number of infected cases and related words in this domain, such as number, death, day, and names of countries, have been repeated more frequently in the real news.

	We also analyze the polarity and subjectivity of real and fake news of the Covid-19 training set. Polarity is a float that lies in the range of [-1,1], where -1 means negative statement and 1 means a positive statement. Objective refers to factual information, whereas subjective sentences generally refer to emotion or judgment and personal opinion. Subjectivity is also a float that lies in the range of [0,1] ~\cite{PolarSub}.  
	
\begin{figure}[H]
		\centering
		\includegraphics[scale=0.5]{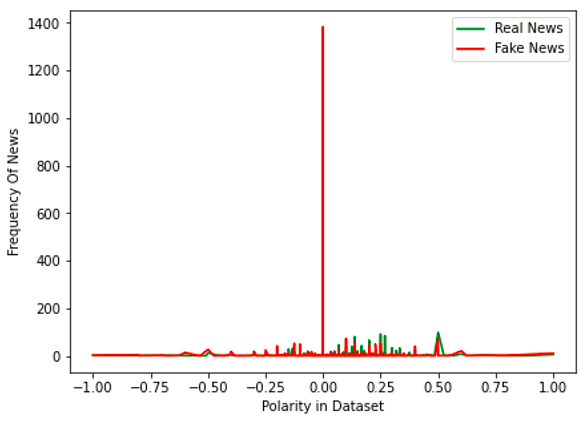}
		\caption{Polarity of real and fake news}
		\label{fig:Polarity}
	\end{figure}

\begin{figure}[H]
		\centering
		\includegraphics[scale=0.55]{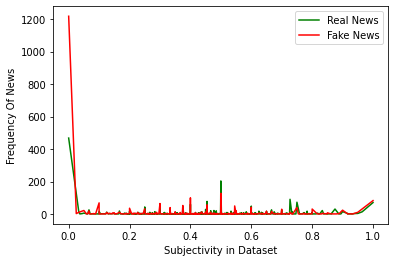}
		\caption{Subjectivity of real and fake news}
		\label{fig:subjective}
	\end{figure}

	Figure \ref{fig:Polarity} and \ref{fig:subjective} show the polarity and subjectivity based on the frequency of real and fake news, respectively. We can see that "zero" is the most frequent type of polarity, and in both real and fake news, positive polarity is more than negative polarity. However, for both classes, negative polarity for fake news is more frequent than for real news, while positive polarity is more pronounced in real news.
	
	In figure \ref{fig:subjective}, we can see in both fake and real news that zero subjectivity is more frequent; but for fake news, it is obviously more. It is also observed that the subjectivity distribution for fake news is higher than that for real news.
	
\begin{figure}[H]
		\centering
		\includegraphics[scale=0.7]{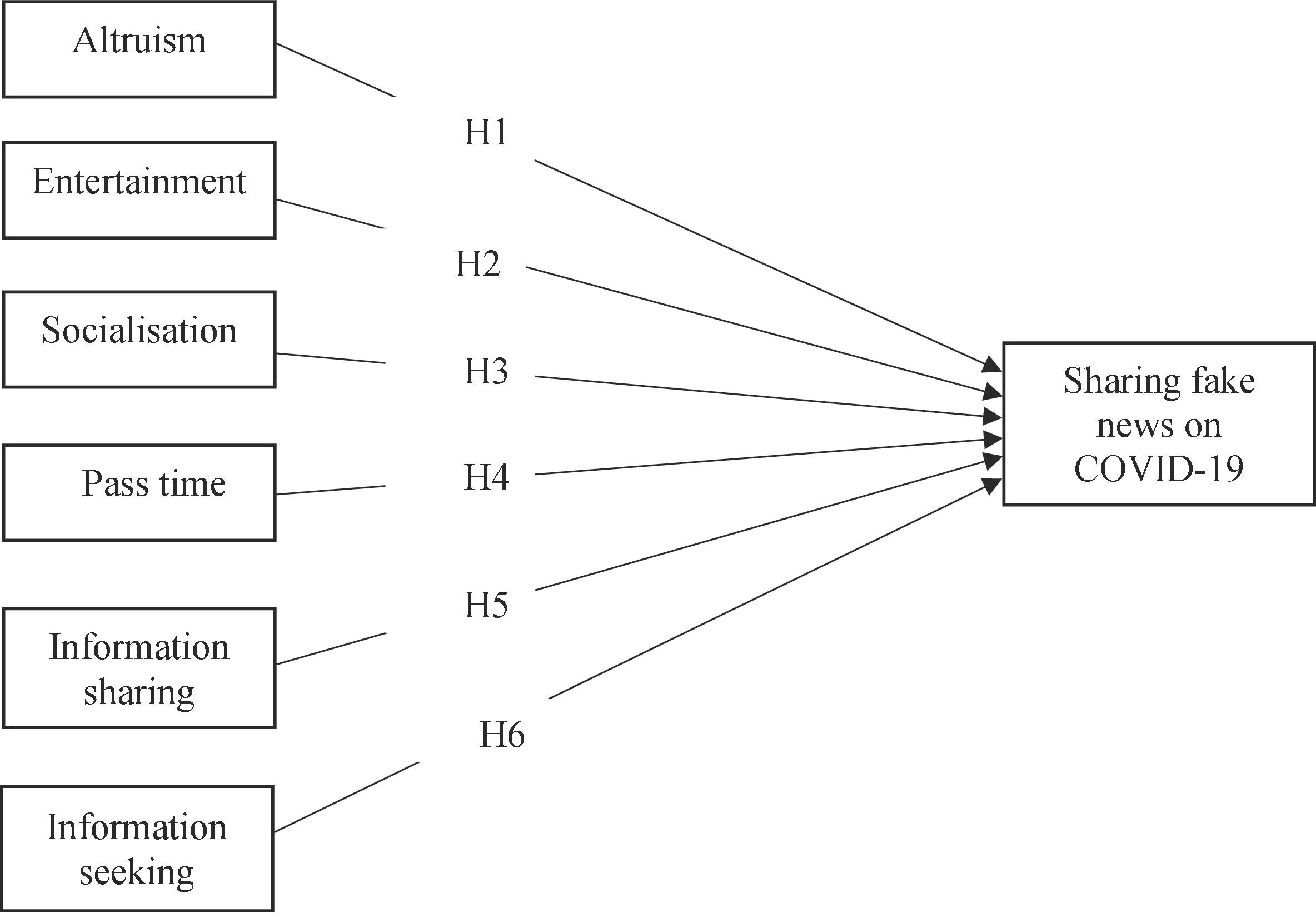}
		\caption{methods of fake news sharing on COVID-19 proposed by \cite{apuke2021fake}.}
		\label{fig:sharing}
	\end{figure}

	Figure \ref{fig:sharing} shows the different methods of fake news sharing with COVID-19 subject on social media proposed by \cite{apuke2021fake}. This study's model was developed with the U\&G theory and previous studies and includes six sharing methods: entertainment, socialization, pass time, altruism, information seeking, and information sharing.
	
	In order to further experiments on the subjectivity of fake news, the fake news with the subjectivity of one in Figure \ref{fig:subjective} is extracted and classified. In Table \ref{tab:sharing}, we can see most of the fake news with high subjectivity shared for information sharing purposes on social media. Also, socialization is another frequent method that is used in fake news sharing. As a result, one of the methods of spreading fake news is sharing interesting information and then using members to re-share that news on social networks.

	\begin{table}[H]
		\centering
		\begin{tabular}{|c|c|}
			\hline
			\textbf{\begin{tabular}[c]{@{}c@{}}Method of \\ Fake new sharing\end{tabular}} & \textbf{Number of news} \\ \hline
			Socialization                                                                 & 16                      \\ \hline
			Alturism                                                                      & 3                       \\ \hline
			Information seeking                                                           & 3                       \\ \hline
			Information sharing                                                           & 49                      \\ \hline
		\end{tabular}
		\caption{Distribution of different sharing methods of fake news on COVID-19}
		\label{tab:sharing}
	\end{table}
	
	%
	%\begin{figure}[H]
	%	\centering
	%	\includegraphics[scale=0.3]{img/Fake_Bi_frequency.png}
	%	\caption{Fake Cloud}
	%	\label{fig:proposed_model}
	%\end{figure}
	
	%\begin{figure}[H]
	%	\centering
	%	\includegraphics[scale=0.3]{img/Fake_Bi_frequency.png}
	%	\caption{Fake Cloud}
	%	\label{fig:proposed_model}
	%\end{figure}

	%\subsection{Classification for Covid-19 dataset}
	%\label{sec:Section 5.2}
	%\subsubsection{Covid-19 dataset}

	\section{Conclusion}
	\label{sec:Section 6}
	
	This paper proposes X-CapsNet for detecting long and short fake news statements. DCNN-CapsNet with margin loss has been proposed for detecting long fake news statements, and MLP-CapsNet with indirect features for short fake news statements. DCNN-CapsNet uses four parallel neural networks. These parallel networks include three different n-grams convolutional layers for feature extraction and a CapsNet layer. MLP-CaCapsNet has been proposed to solve the problem of short fake news statements that, in addition to the representation layer, we use an indirect features vector created by concatenating news speaker profile information and sentiment, polarity, and sentence information of a fake news article. Different pre-trained representation models and different iterations of the dynamic routing algorithm for the CapsNet have been used to evaluate the proposed models. Finally, models have been tested on two recent well-known datasets in the field, namely the Covid-19  with long fake news statements and the Liar datasets as a dataset with short news statements. Our result shows that using these models can improve the performance of state-of-the-art baselines. 
	\ifCLASSOPTIONcaptionsoff
	\newpage
	\fi

	
	
	%\bibliographystyle{IEEEtran}
	
	\bibliographystyle{IEEEtran}
	\bibliography{Fake_news_arxiv}
	%\bibliographystyle{plain}
	%\begin{thebibliography}{1}
	%\end{thebibliography}
	\newpage	

\end{document}